\documentclass[runningheads]{llncs}
\usepackage[T1]{fontenc}
\usepackage{caption}
\usepackage{graphicx}
\usepackage{xcolor}
\usepackage{soul}
\usepackage{fancyhdr}
\usepackage{url}

\begin{document}

\newcommand{\acronym}{GNN-XAR}
\title{\acronym: A Graph Neural Network for
Explainable Activity Recognition in Smart Homes}
\titlerunning{Explainable GNN for Smart Home HAR}
%
\author{
Michele Fiori \and
Davide Mor\and
Gabriele Civitarese\and 
Claudio Bettini}
\authorrunning{Fiori et al.}
%
\institute{University of Milan, Milan, Italy \\
\email{\{michele.fiori, gabriele.civitarese, claudio.bettini\}@unimi.it}\\ \email{d.mor1@campus.unimib.it}}
\maketitle              

\begin{abstract}
Sensor-based Human Activity Recognition (HAR) in smart home environments is crucial for several applications, especially in the healthcare domain. The majority of the existing approaches leverage deep learning models. While these approaches are effective, the rationale behind their outputs is opaque. Recently, eXplainable Artificial Intelligence (XAI) approaches emerged to provide intuitive explanations to the output of HAR models. To the best of our knowledge, these approaches leverage classic deep models like CNNs or RNNs. Recently, Graph Neural Networks (GNNs) proved to be effective for sensor-based HAR. However, existing approaches are not designed with explainability in mind. In this work, we propose the first explainable Graph Neural Network explicitly designed for smart home HAR.
Our results on two public datasets show that this approach provides better explanations than state-of-the-art methods while also slightly improving the recognition rate.

\end{abstract}

{\small\textcolor{red}{\textit{This manuscript has been peer-reviewed and published, please cite the published version: \url{https://doi.org/10.1007/978-3-032-10554-7_19}}}

\section{Introduction}

The recognition of Activities of Daily Living (ADLs) in Smart Home environments is a widely studied research topic in the pervasive computing community~\cite{bouchabou2021survey}. Recognizing the daily activities that humans do in their daily life at  home (e.g., cooking, watering plants, taking medicines) has several important healthcare applications, including the early detection of cognitive decline~\cite{riboni2016smartfaber}.

The majority of the approaches in the literature are based on deep learning models, mainly due to their effectiveness in reaching high recognition rates~\cite{gu2021survey}. The most common architectures used for ADLs recognition are Convolutional~\cite{arrotta2022dexar} and Recurrent~\cite{liciotti2020sequential} neural networks. However, these approaches may not fully capture the spatiotemporal properties of sensor data. In the literature, Graph Neural Networks (GNNs) have emerged as a promising approach for time series classification~\cite{duan2022multivariate}. In GNNs, sensor data time windows are encoded as graphs capturing both spatial and temporal relationships between sensors. While the majority of existing studies focused on human activity recognition with mobile/wearable devices~\cite{mondal2020new}, only a few GNN-based approaches have been proposed for ADLs recognition in smart home environments~\cite{ye2023graph,s24123944}. 

In general, deep learning models are often considered as ``\textit{black boxes}'' mapping windows of sensor data into activities, and it is challenging to understand the rationale behind their decisions.
The field of eXplainable Artificial Intelligence (XAI) has the goal of mitigating this problem by providing human-understandable explanations to the output of machine learning models~\cite{arrieta2020explainable}. 

Since important decisions in ambient assisted living applications may rely on the output of ADLs recognition, inferring \emph{why} a specific ADL was predicted by the classifier is crucial to provide understandable, trusted, and transparent solutions~\cite{wolf2019explainability}. For instance, XAI would allow clinicians to increase their trust in decision support systems that rely on ADLs recognition (e.g., supporting early detection of cognitive decline).
Data scientists may also benefit from explanations to refine the recognition system, the sensing infrastructure, or the training set. 

A few works proposed explainable ADLs recognition in smart homes environments~\cite{arrotta2022dexar,das2023explainable}. However, to the best of our knowledge,  eXplainable GNN approaches for sensor-based ADLs recognition in smart homes have not been explored yet. Therefore, in this paper we present \acronym{}, the first explainable GNN-based system for ADLs recognition. Specifically, \acronym{} dynamically constructs a graph starting from windows of environmental sensor data taking into account spatial and temporal aspects. Each graph is processed by a Graph Convolutional Network (GCN) for ADLs classification. An adapted state-of-the-art XAI method specifically designed for GNNs is in charge of determining the most important nodes and arcs of the input for activity classification. This information is finally used to generate an explanation in natural language.

To sum up, the contributions of this paper are the following:

\begin{itemize}
    \item We propose \acronym{}: the first Explainable Graph Neural Network system for Smart Home HAR.
    \item Starting from windows of raw sensor data, \acronym{} dynamically constructs a graph that a GNN processes to classify the most likely activity.
    \item For each prediction, \acronym{} leverages an eXplainable AI approach to produce explanations in natural language. 
    \item Our results show that \acronym{} generates superior explanations with respect to state-of-the-art explainable HAR methods, while slightly improving the recognition rate.
\end{itemize}

\section{Related Work}

\subsection{GNN-based methods for HAR}
GNNs have been widely adopted in IoT scenarios, in applications including  multi-agent interaction, Human State-dynamic, sensor interconnection, and autonomous vehicles~\cite{10.1145/3565973}.
Considering sensor-based human activity recognition, GNNs have been mainly proposed to recognize activities from mobile/wearable devices~\cite{10174597}.

The differences between existing work lies in how the graph is constructed from sensor data.
For instance, a common solution is to consider each time window as a node with the goal of performing node classification~\cite{9680185,9767342,9874212}.
Other works consider a node for each sensor, considering a graph classification task~\cite{9995660}. 

Only a few works applied GNNs to ADLs recognition in smart home environments. For example, \cite{s24123944} uses graphs to model sensor dependencies. In that work, a graph is built such that each node represents an environmental sensor, while a directed arc represents the influence of the behavior of one sensor to another one. The arcs of the graph structure and their weights are learned using an attention mechanism. The graph classification task is then considered to perform ADLs classification.
The work in~\cite{ye2023graph} is closely related to \acronym{} since it dynamically constructs each graph from the time window of sensor data, where each node is a sensor event. Differently from our work, the arcs are automatically learned from the network considering spatial and temporal properties at the same time.  

While it may be possible to apply XAI techniques on such methods, it would be challenging to obtain meaningful explanations. For instance, considering the work in~\cite{s24123944}, it may be possible to obtain explanations only about the sensors triggered consecutively, without considering longer temporal relationships. On the other hand, since in~\cite{ye2023graph} the arcs are automatically learned, they are not associated with a specific semantic and hence it would be challenging to explain them.

\subsection{XAI methods for Human Activity Recognition}

The first attempts for explainable sensor-based activity recognition considered simple inherently interpretable models~\cite{bettini2021explainable,atzmueller2018explicative,guesgen2020using,khodabandehloo2021healthxai}. However, such models usually underperform deep learning models that, on the contrary, are more complex to explain.

A few works proposed XAI methods for deep learning for sensor-based activity recognition~\cite{jeyakumar2023x,meena2023explainable}. However, only a few of them focused on ADLs recognition in smart homes.
For instance, DeXAR~\cite{arrotta2022dexar} leverages CNN models and explores the use of various XAI methods for computer vision by converting sensor data into semantic images.
Similarly, the work in~\cite{das2023explainable} applies several XAI methods to LSTM-based neural networks. Both works generate explanations in natural language for non-expert users.

While eXplainable Graph Neural Networks have been studied in the general machine learning community~\cite{agarwal2023evaluating}, this is the first work exploring this combination for sensor-based ADLs recognition in smart homes.

\section{\acronym{} under the hood}

In this work, we consider a smart home environment equipped with binary environmental sensors (e.g., motion sensors, magnetic sensors, pressure sensors). We assume that the smart-home is inhabited only by a single resident, and that sensor events are the result of the interaction of the subject with the environment. We also assume that the same timestamp will never be assigned to two distinct sensor events, hence we consider a total order in the sensor event sequence \footnote{Note that this is a realistic assumption, since typically events are queued by a single process on the gateway that assigns different timestamps even if the events occurred at the same time, without any impact on our results.}. The goal of \acronym{} is to provide the most likely activity from time windows of sensor events and, at the same time, to provide an explanation in natural language about the aspects of the input that mostly contributed to the prediction.

\subsection{Overall Architecture}
Figure~\ref{fig:architecture} depicts the architecture of \acronym{}.

   \begin{figure}[h!]
        \centering
        \includegraphics[width=\textwidth]{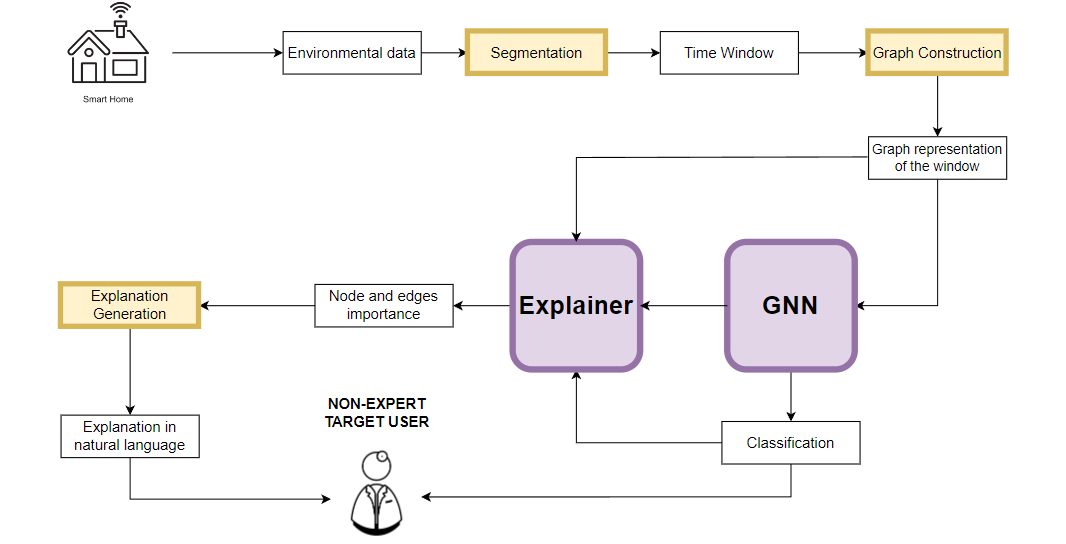}
        \caption{Overall architecture of \acronym{}.}
        \label{fig:architecture}
    \end{figure}

First, the stream of environmental sensor data is segmented into fixed size overlapping time windows. Each time window is then processed by the \textsc{Graph Construction} module to obtain a graph representation of the window, encoding both spatial and temporal properties with an heuristic-based approach. Each graph is processed by the \textsc{GNN} module for ADLs classification. Then, the \textsc{Explainer} module leverages posthoc XAI methods to obtain the nodes and arcs that were the most important in the input to obtain the classified activity. Posthoc XAI approaches consider the model as a black box and generate explanations by analyzing the relationships between inputs and outputs. Finally, the \textsc{Explanation Generation} module uses this information to generate an explanation in natural language for non-expert users.

\subsection{Graph Construction}

The \textsc{Graph Construction} module dynamically constructs a graph $G_w$ starting from a temporal window $w$ of sensor events. \acronym{} leverages a heuristic-based strategy for graph construction, taking into account spatial and temporal properties (that will be leveraged to generate meaningful explanations).

Let $w=\langle E_1, E_2, \dots, E_n \rangle$ be a temporal window including $n$ sequential sensor events. An event $E_i$ is associated with the following information:
\begin{itemize}
    \item An identifier of the sensor that produced it ($E_i^{id}$), that encodes the sensor type (magnetic, motion ...) and the position (fridge, sofa...).
    \item The event type ON or OFF ($E_i^{type}$).
    \item The timestamp of the event ($E_i^{ts}$).
\end{itemize}

In our framework, we consider the events differently based on the type of sensor that generated them:
\begin{itemize}
    \item The first type of sensor includes sensors whose both activation and deactivation require explicit actions by the user (e.g. opening or closing a cabinet). In this case, we are interested in both the activation and deactivation (i.e. ON and OFF) events.
    \item The second type includes sensors that are automatically deactivated after some time, for instance, motion sensors. In this case, we are interested in the ON event and in the duration of the \textit{active state} of the sensor, computed as the time from the ON event to the OFF event.\footnote{In the case of motion sensors if there are multiple subsequent detections of movements, a single active state is considered. This is also how the events are reported in the public datasets we considered.}
\end{itemize}

Given a window $w$, we denote the corresponding graph with $G_w = (V,A)$, where $V$ is the set of nodes and $A$ is the set of arcs. In \acronym{}, the set $V$ is created as follows:

\begin{itemize}
    \item We add a node $v$ for each event in $w$ (activation or deactivation) generated by the first type of sensor. These are \textbf{event nodes}.
    \item We add a node $v$ for every \textit{active} state of a sensor of the second type. These are \textbf{state nodes}.
\end{itemize}
In our system, each node has the sensor identifier as a feature. As common in deep learning, a (trainable) embedding layer computes an embedding vector representing this feature. State nodes have the duration of the corresponding \textit{active state} as an additional feature.

The set of arcs $A$ is created as follows:
For each pair of node $v_i,v_j \in V$ such that $i \neq j$, there is an oriented arc $(v_i,v_j)\in A$ in the following two cases:
\begin{enumerate}
    \item $v_i$ and $v_j$ are event nodes derived from consecutive events generated by the same sensor $S$ (i.e., there are no other events from $S$ between them). This is shown in Figure~\ref{fig:XT1}.
    \begin{figure}
        \centering
        \includegraphics[width=0.7\textwidth]{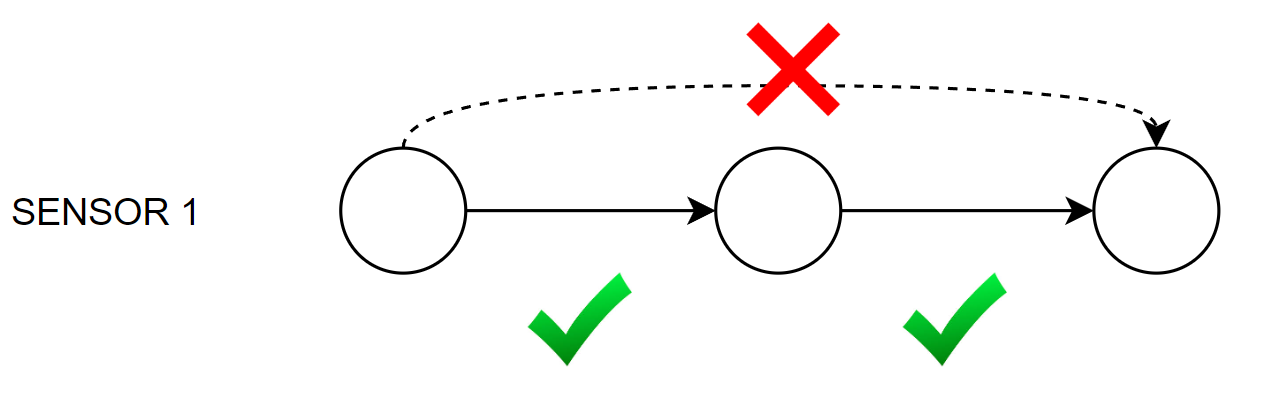}
        \caption{Arcs between event nodes generated by the same sensor.}
        \label{fig:XT1}
    \end{figure}
    \item $v_i$ and $v_j$ are state nodes derived from consecutive active states of the same sensor $S$. 
    
    \item $v_i$ and $v_j$ are derived from events/states generated by different sensors $S_a$ and $S_b$ and there are no other events/states generated by $S_a$ or $S_b$ between them. This temporal relationship is computed considering as timestamp of an active state the timestamp of the activation (ON) event. 
    This is shown in Figure~\ref{fig:XT2}.

        \begin{figure}
        \centering
        \includegraphics[width=0.7\textwidth]{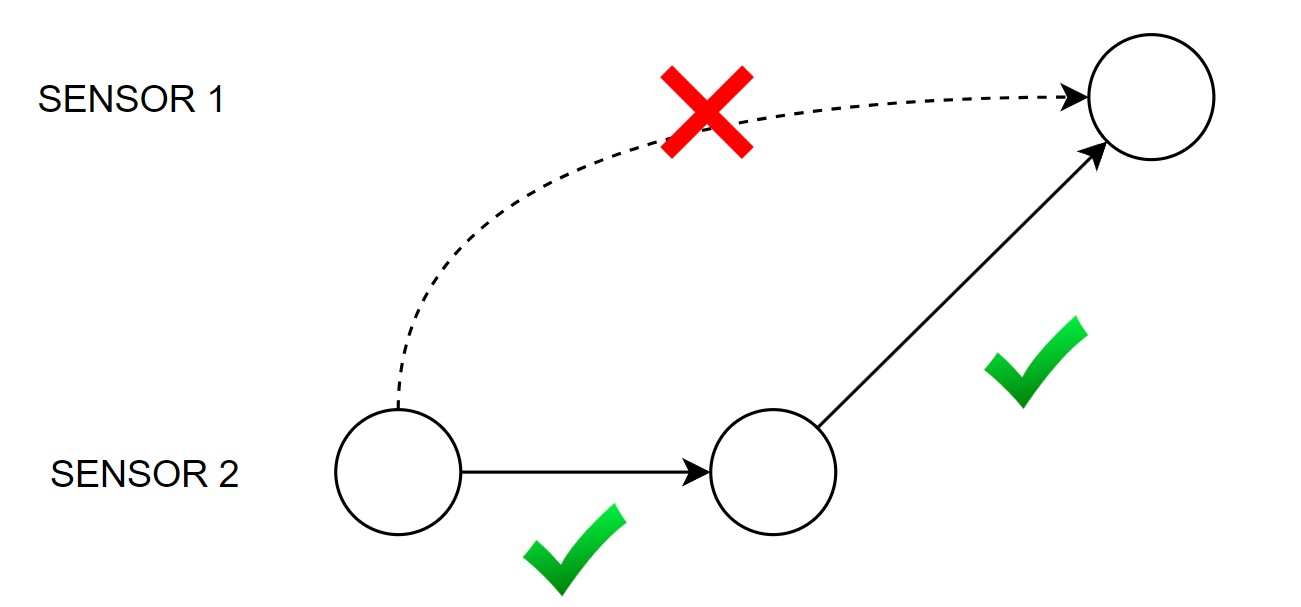}
        \caption{Arcs between event or state nodes generated by different sensors.}
        \label{fig:XT2}
    \end{figure}
\end{enumerate}

Each arc $(v_i,v_j)\in A$ has as associated feature the time difference between the timestamp of the event/state corresponding to $v_j$ and the timestamp of the event/state corresponding to $v_i$, considering as timestamp of an active state the timestamp of the activation (ON) event.

This method for building the graph has also the advantage of maintaining a sparse structure, avoiding the graph degeneration into a fully connected graph.

Our graph construction strategy makes it possible to consider spatiotemporal relationships between sensor events. Considering temporal aspects, a directed arc from a node $v_i$ to a node $v_j$ models the fact that the event/state corresponding to $v_i$ occurred before the event/state corresponding to $v_j$. This temporal relationship is also quantitative since the arc feature encodes the time distance between them. Moreover, when $v_i$ and $v_j$ are generated by different sensors $S_a$ and $S_b$, the directed arc also implicitly encodes spatial relationships about the resident interacting with sensors in different home positions.

A challenge with graphs with a variable number of nodes is that it complicates the graph pooling process (i.e., creating an embedding for the whole graph).
For instance, pooling by concatenation would lead to graph embeddings of different shapes. 
\acronym{} solves this problem by augmenting $V$ with a fixed number of \textit{super-nodes}: fictitious nodes not corresponding to real sensor events or states. 
Specifically, we add a super node $SN_S$ to $V$ for each sensor $S$. We also add an arc $(v_i, SN_S)\in \mathbf{A}$ if $v_i$ corresponds to an event/state generated by the sensor $S$. Note that, if a window $w$ does not contain events/states generated by a sensor $S_a$, $V$ would include the super node $SN_{S_a}$ without associated arcs.
Hence, all the nodes corresponding to an event/state generated by a sensor are connected to its super-node. 
Super-nodes have the role of summarizing all the graph information into a fixed number of nodes. Thanks to this approach, it is possible to concatenate the information of the whole graph only by performing pooling on super-nodes.

In the following, we show a simple example of how to build a graph $G_w$ starting from a window $w$ including active states from three sensors.
The window $w$ and its states are depicted in Figure~\ref{fig:EX1}.

\begin{figure}
    \centering
    \includegraphics[width=1\textwidth]{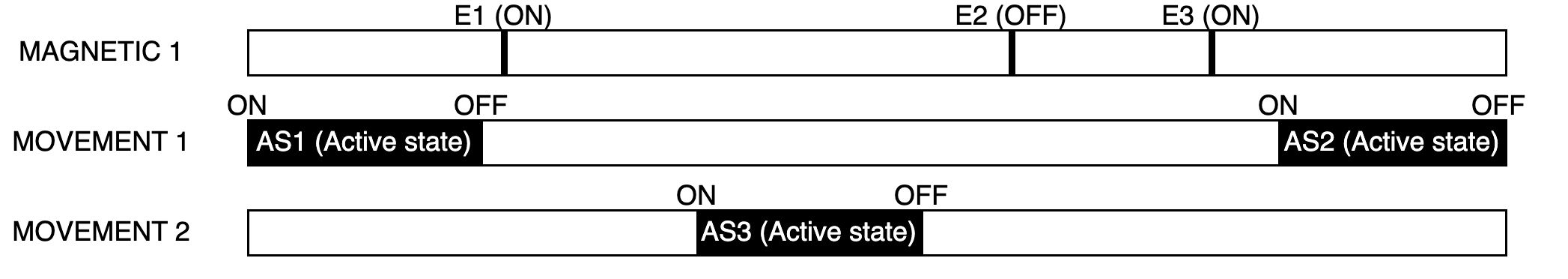}
    \caption[Graph construction example 1]{Example of a time window. For the magnetic sensor, the black lines represent the events. For the movement sensors, the black regions represent the active states}
    \label{fig:EX1}
\end{figure}

Figure~\ref{fig:EX2} shows the resulting graph $G_w$, that is constructed with the following steps:
\begin{itemize}
    \item For each sensor $S$, a super-node is created (represented as squares in the figure).
    \item For each event/state a node $v_i$ is generated (represented as circles in the figure).
    \item Each event/state generated by $S$ is connected to the respective super-node associated with $S$ (the dashed arrows).
    \item We include arcs based on the rules defined above (the solid arrows).
    
\end{itemize}

\begin{figure}
    \centering
    \includegraphics[width=1\textwidth]{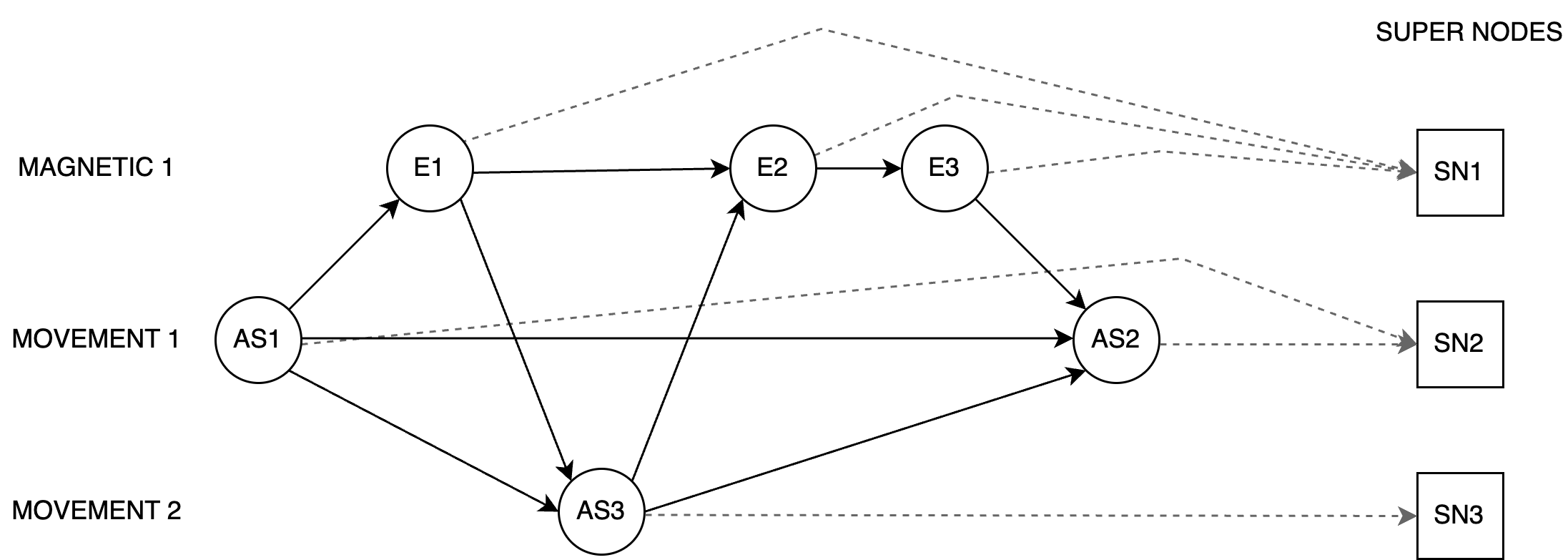}
    \caption[Graph construction example 2]{The directed graph computed from the sensor window in Figure~\ref{fig:EX1}. } 

    \label{fig:EX2}
\end{figure}

\subsection{Graph Neural Network}

In the following, we describe the steps for graph classification.

\subsubsection{Message Passing}

Message passing is a crucial step in GNNs for transmitting information between nodes in a graph. This technique allows nodes to share and update their features based on the features of their neighbors, facilitating the extraction of meaningful patterns and relationships in graph-structured data. 

Information propagates through the graph in two distinct steps. The first step aggregates and processes all the known information, including event duration and distances between events, to compute the new node features, while the second step has the goal of further spreading this information into the graph. The second step is particularly useful for spreading information to the super nodes.

More specifically, in the first iteration, for each node, a message is computed by applying a linear layer to the concatenation of the node embeddings and the arc features. This linear layer reduces the message's dimension to match the original node feature dimension. Subsequently, the aggregate message for each node is computed using a sum aggregation function, and the new node feature is obtained by summing the previous embedding with the aggregated message.

The second phase involves a simplified propagation in which only the new node embeddings are considered. Given that less information is processed in this phase, no linear layer is applied. Instead, a sum aggregation function is used directly. The same update function from the first phase is then applied, summing the previous node feature with the aggregated messages to compute the new node feature.

Although this propagation process can be iterated multiple times, the high connectivity of the graphs generated by \acronym{} makes it possible to leverage a small number of iterations to allow information to travel throughout the entire graph.


\subsubsection{Graph Pooling and Classification}

Figure~\ref{fig:GNN} shows the GNN model architecture of \acronym{}.
Graph Pooling consists in generating an embedding that is representative of the whole graph. As we previously mentioned, since the number of nodes is different for each instance, we leverage super-nodes. Thus, the pooling strategy proposed for this model consists of considering only the embeddings of super nodes and concatenating them to obtain a vector of length equal to the embedding dimension multiplied by the number of sensors.

The classification is carried out using linear layers: the flattened embedding encoding the graph is passed through two linear layers, each one followed by a LeakyReLU function. Finally, the output of the network is the probability distribution over the possible ADLs, obtained thanks to a softmax layer.

   \begin{figure}[h!]
        \centering
        \includegraphics[width=\textwidth]{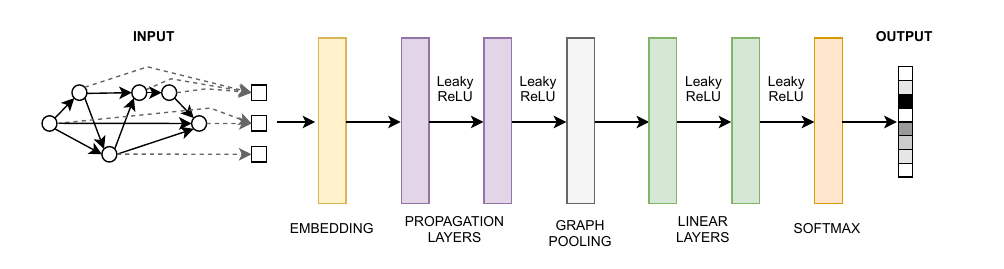}
        \caption{The model architecture of \acronym{}.}
        \label{fig:GNN}
    \end{figure}

\subsection{Explainer}

XAI methods applied on GNNs aim to find the subset of nodes and arc that mostly contributed to a specific prediction. In \acronym{}, a node is selected for the explanation if the corresponding sensor event/state was important for classifying the activity; an arc is selected if the specific order of sensor events/states was important for classifying the activity.

\acronym{} leverages the GNNexplainer~\cite{ying2019gnnexplainer} method for explanations. Through an optimization method, GNNexplainer derives a subgraph maximizing the mutual information between the GNN's prediction and the prediction that would have been obtained by the GNN based only on this subgraph. This is achieved by perturbing the graph and its features, and observing the effect of these perturbations on the GNN's predictions. Given the most likely ADL predicted by the GNN, the input graph and the GNN model, GNNexplainer computes importance values for nodes and arcs. The algorithm leverages gradient descent to derive node and arc masks that modulate the information spread in the graph during the message-passing procedure.

In its original version, the GNNexplainer algorithm includes the multiplication of each node feature by a value in the range $[0,1]$ to generate the node mask.
However, this approach cannot be adopted in GNN-XAR since the sensor id node feature represents a categorical value, 
the perturbation of that value would not convey the intended meaning. For this reason, we apply the original GNNExplainer but extract only the arc mask. We then compute the importance of each node as the importance of the arc connecting it to its corresponding super node. 
The intuition behind this strategy is that there is only one arc through which the information from the node is propagated to the super node. Since it is the super node that is used for the classification, the importance of this arc is a good indicator of the importance of the information conveyed by the node.

\begin{figure}
    \centering
    \includegraphics[width=0.8\textwidth]{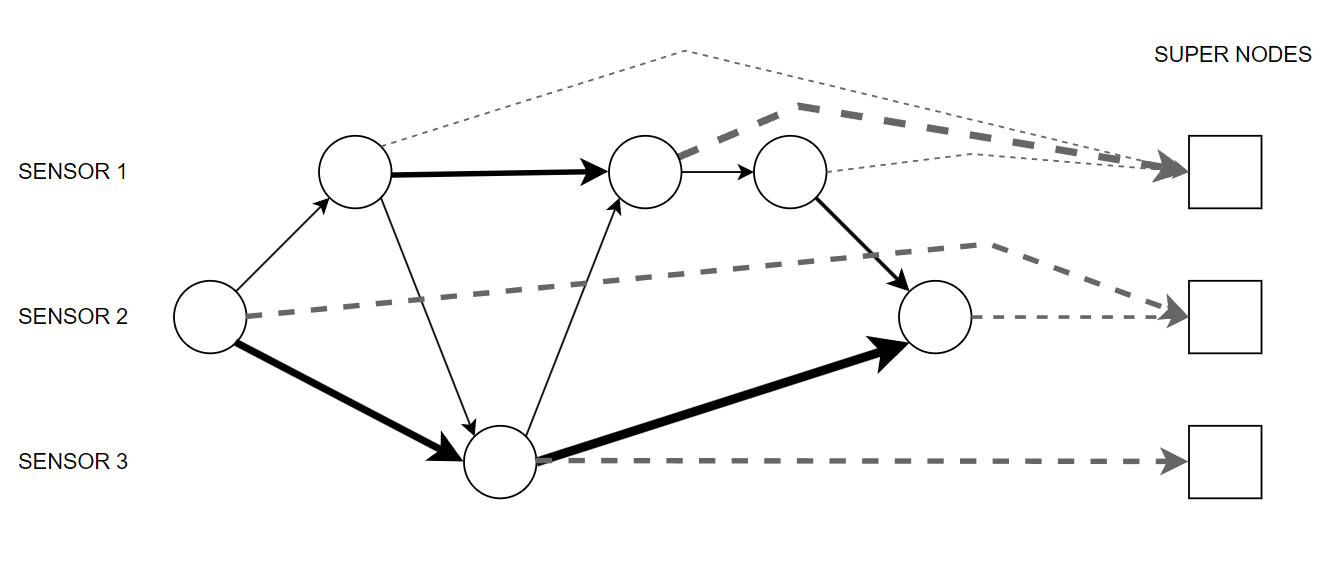}
    \caption[How explainer algorithm works]{Output of the original GNNExplainer limited to arcs importance. The thickness of each arrow represents the importance values on the arcs.}
    \label{fig:MSK-1}
\end{figure}

Figure ~\ref{fig:MSK-1} represents the arcs importance as computed by the original GNNExplainer on an example in our domain. Note that the dashed lines, connecting nodes with super nodes, also have different thicknesses representing different importance.


As stated above, our adapted version of GNNexplainer assigns as importance of each node the importance of the arc connecting it with its super node.
Figure ~\ref{fig:EX4} shows the output of our adapted version of GNNexplainer on the same example of Figure \ref{fig:MSK-1}. Note that super nodes are not part of the output and that an importance value is associated with each node, denoted in the figure by the thickness.

\begin{figure}
    \centering
    \includegraphics[width=0.65\textwidth]{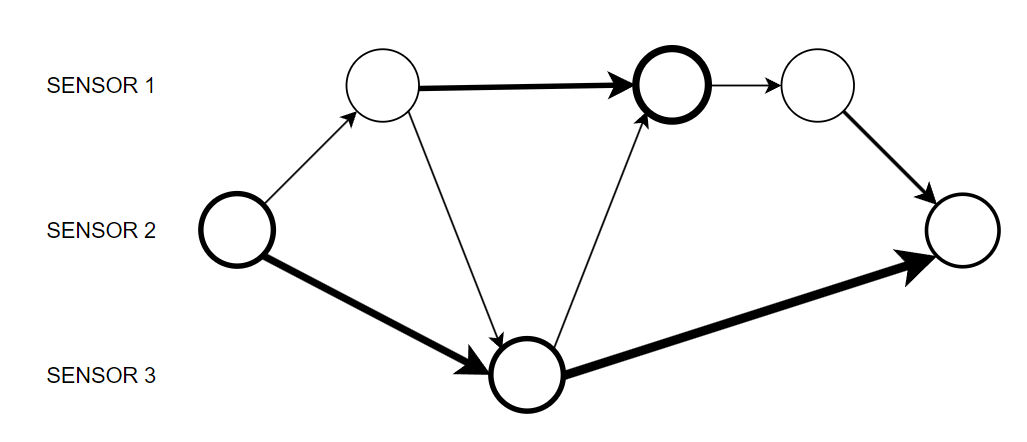}
    \caption[How to obtain a node mask]{Output of our adapted version of GNNExplainer on the same input example of Figure \ref{fig:MSK-1}.}
    \label{fig:EX4}
\end{figure}

GNNexplainer is a non-deterministic algorithm. For this reason, it requires multiple executions, and only the average of the masks obtained at each execution is considered for the explanations.
A major challenge of using GNNexplainer is that the importance values obtained for nodes and arcs are not directly comparable, since they range in different intervals. 
Since we observed that arcs scorers are usually associated with lower importance values, we rescale them by a multiplicative factor such that the importance values of the nodes and the ones for the arcs have the same mean.

Finally, \acronym{} aims at extracting a supgraph $G_w^\star = (V^\star,A^\star)$ where $V^\star$ is the set of the most important nodes, while $A^\star$ is the set of the most important arcs. A straightforward approach to achieve this task would be using a threshold on the importance value.
However, we observed that different predictions are usually associated with importance values in completely different ranges. Hence, defining a robust threshold is challenging. We mitigated this problem by adopting a clustering approach. Indeed, we cluster arcs and nodes based on their importance values.
We consider as the most important arcs and nodes the ones in the cluster associated with the highest importance values.




\subsection{Generating explanations in natural language}

As a final step, \acronym{} converts $G^\star$ into a natural language explanation for non-expert users.

Given the set of most important arcs $A^\star$, we compute the longest path.
We then use a heuristic-based approach similar to the one proposed in~\cite{arrotta2022dexar} to generate from this path a natural language explanation.
For instance, the continuous activation of certain sensors implies that the resident has moved toward the sensor multiple times: in this case, in the path explanation, the expression ``\textit{multiple times}'' is added:\\

  \textit{``I predicted preparing a meal mainly due to the following observations: Bob was near the fridge, then he opened the fridge multiple times''}\\



\section{Experimental Evaluation}

\subsection{Datasets}

Two datasets have been used to evaluate the model proposed in this work. They are \textit{CASAS Milan}~\cite{cook2009assessing} and \textit{CASAS Aruba}~\cite{cook2010learning}. 

\subsubsection{CASAS Milan}    

CASAS Milan consists of data gathered from the home of a female adult volunteer living with a pet and where the woman's children visited periodically. The dataset contains about three months of recording and includes the following classes: Bed-to-Toilet, Chores, Desk Activity, Dining room Activity, Evening Medications, Guest Bathroom, Kitchen Activity, Leave Home, Master Bathroom, Meditate, Watch TV, Sleep, Read, Morning Medications, Master Bedroom Activity. Some of these activities are very under-represented in the dataset; for this reason, following what has been done in~\cite{arrotta2022dexar}, the less represented classes, that are Bed to Toilet, Chores, Meditation, Evening Medications and Morning Medications, are not taken into consideration. Moreover, Master Bathroom and Guest Bathroom have been fused obtaining a new class that contains similar activities in a number comparable with the other classes.

\subsubsection{CASAS Aruba} 
CASAS Aruba is a dataset collected in the home of a woman whose children and grandchildren visited regularly. The dataset contains Meal Preparation, Relax, Eating, Work, Sleeping, Wash Dishes, Bed to Toilet, Enter Home, Leave Home, Housekeeping and Resperate. Resperate and Bed To toilet classes have been dropped according to what has been done in~\cite{liciotti2020sequential}.

\subsection{Implementation details}
We implemented \acronym{} using Python 3.10.5, using Pytorch and Pytorch Geometrics for the models and the explainer.
Other libraries used include Scikit-learn for the evaluation, Networkx for graph visualization, and Pandas and Numpy for data processing.

\subsection{Evaluation}

\subsubsection{Baseline}
We decided to focus our comparison of \acronym{} only with state-of-the-art explainable ADL recognition methods, selecting the one that demonstrated the highest recognition accuracy in the literature.
For this reason, we chose DeXAR~\cite{arrotta2022dexar} as a baseline, since it is the method that meets these criteria. DeXAR converts sensor data into semantic images, leveraging XAI methods for computer vision to generate natural language explanations. Since \acronym{} uses a posthoc explanation method, we compare DeXAR when used with LIME~\cite{ribeiro2016should}.

The original DeXAR implementation also considered previously predicted ADLs as input, while this aspect is not captured by \acronym{}. Hence, we implemented a version of DeXAR not considering past activities.

\subsubsection{Experimental setup}

We consider a standard $70\%$-$20\%$-$10\%$ split to partition the datasets into training, test and validation sets.
The models have been trained using the early stopping strategy with a patience of $50$ epochs, the Adam optimizer with a learning rate of $0.0001$, and a CrossEntropy loss function.

For segmentation, we used the same hyper-parameters suggested in~\cite{arrotta2022dexar} for the CASAS datasets, where the window size is $360$ seconds with an overlap factor of $80\%$.
We discarded all the windows not corresponding to an activity label (i.e., transitions or other activities) as well as temporal windows without sensor events.



\subsection{Evaluation Metrics}

We use the standard metrics for precision, recall, and F1 score to assess the recognition rate of \acronym{}. 
These metrics provide a comprehensive understanding of the model's performance from different perspectives.


However, evaluating the effectiveness of explanations is more challenging. A standard way adopted in the literature involves user surveys~\cite{arrotta2022dexar,das2023explainable,jeyakumar2023x}. However, such method is time and money-consuming.
We leverage a recent work proposing LLMs to automatically compare alternative XAI methods, since it proved to be aligned with user surveys~\cite{fiori2024using}.
Specifically, we provide to an LLM the explanations generated by \acronym{} and DeXAR on the same window, asking the LLM to choose the best one (using the prompt proposed in~\cite{fiori2024using}).


\subsection{Results}

\subsubsection{Classification results}

Tables~\ref{tab:clf_milan} and ~\ref{tab:clf_aruba} compare \acronym{} and DeXAR considering the F1 score for each class.
We observe that our approach achieves slightly better recognition rates in the overall F1 score for both datasets.
By observing the confusion matrices in Figure~\ref{fig:cm_milan} and \ref{fig:cm_aruba}, both models struggle to distinguish activities taking place in the same room, like Bathroom and Dress Undress. 

Considering the CASAS Milan dataset, this is likely due to the fact that the wardrobe is located in the master bedroom near the bathroom entrance. To distinguish between these two activities, it is probably necessary to consider additional context information, such as past activities and time. Another remarkable difference between the two models is the higher f1 score of \acronym{} on Leave Home (see table~\ref{tab:clf_milan}). In fact, this ADL strongly depends the temporal order of sensor events, that is better captured by our GNN model.

\begin{figure*}
  \centering
  \begin{tabular}{c @{\qquad} c }
    \includegraphics[width=.46\linewidth]{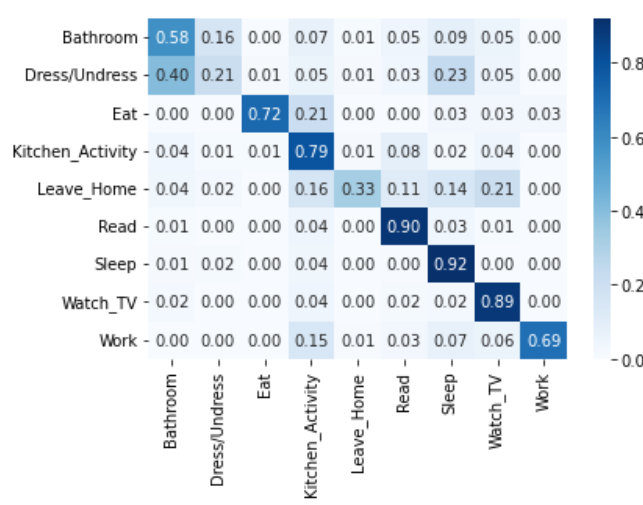} &
    \includegraphics[width=.46\linewidth]{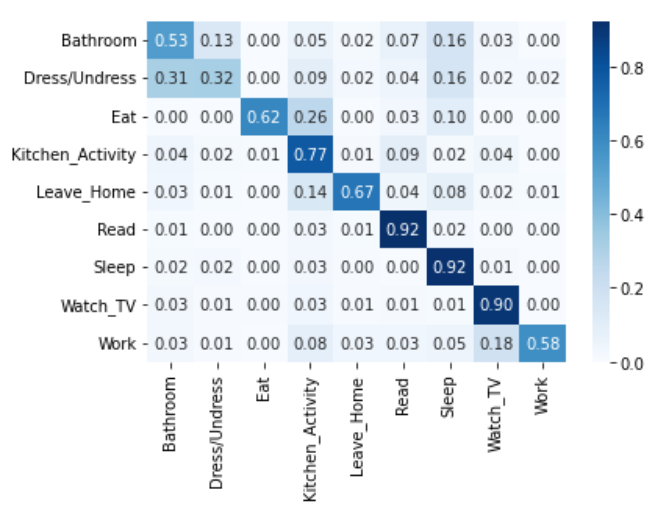} \\
    \small DeXAR & \small \acronym{}
  \end{tabular}
  \caption[Classification results on CASAS Milan]{Confusion Matrices for CASAS Milan.}
  \label{fig:cm_milan}
\end{figure*}

Considering the CASAS Aruba dataset, the GNN model performs better than DeXAR for almost all the activities. The main difference with respect to the results obtained in CASAS Milan is that two activities are completely misclassified by both \acronym{} and DeXAR: washing dishes and housekeeping. These two activities are the least represented in the dataset. Wash dishes is always confused with meal preparation. Similarly to CASAS Milan,  the activities that benefit more from the GNN model are entering home and leaving home.

\begin{table}
\begin{center}
\begin{tabular}{ccc}
\hline
& DeXAR~\cite{arrotta2022dexar} & \acronym{} \\
\hline
Bathroom & \textbf{0.55} & 0.53 \\
Dress/Undress & 0.26 & \textbf{0.37} \\
Eat & \textbf{0.67} & 0.61 \\
Kitchen activity & \textbf{0.77} & \textbf{0.77} \\
Leave Home & 0.46 & \textbf{0.74} \\
Read & 0.90 & \textbf{0.91} \\
Sleep & 0.85 & \textbf{0.87} \\
Watch TV & 0.84 & \textbf{0.89} \\
Work & \textbf{0.80} & 0.70 \\
\hline
weighted avg. & 0.77 & \textbf{0.81}\\
\hline
\hspace{5pt}
\end{tabular}
\caption{CASAS Milan: Classification results (F1 score).}
\label{tab:clf_milan}
\end{center}
\end{table}

\begin{table}
\begin{center}
\begin{tabular}{ccc}
\hline
& DeXAR~\cite{arrotta2022dexar} & \acronym{} \\
\hline
Eating & 0.69 & \textbf{0.75} \\
Enter Home & 0.53 & \textbf{0.76} \\
Housekeeping & 0.09 & \textbf{0.14} \\
Leave Home & 0.71 & \textbf{0.82} \\
Meal Preparation & 0.80 & \textbf{0.81} \\
Relax & 0.94 & \textbf{0.96} \\
Sleeping & 0.93 & \textbf{0.96} \\
Wash Dishes & \textbf{0.06} & 0.00 \\
Work & 0.79 & \textbf{0.87} \\
\hline
weighted avg. & 0.90 & \textbf{0.92}\\
\hline
\hspace{5pt}
\end{tabular}
\caption{CASAS Aruba: Classification results (F1 score).}
\label{tab:clf_aruba}
\end{center}
\end{table}

\begin{figure*}
  \centering
  \begin{tabular}{c @{\qquad} c }
    \includegraphics[width=.46\linewidth]{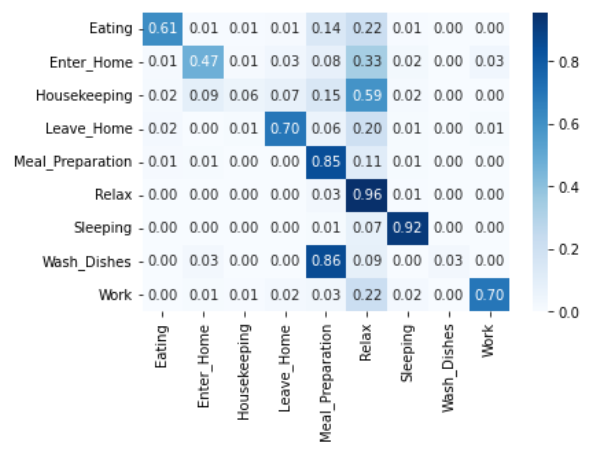} &
    \includegraphics[width=.46\linewidth]{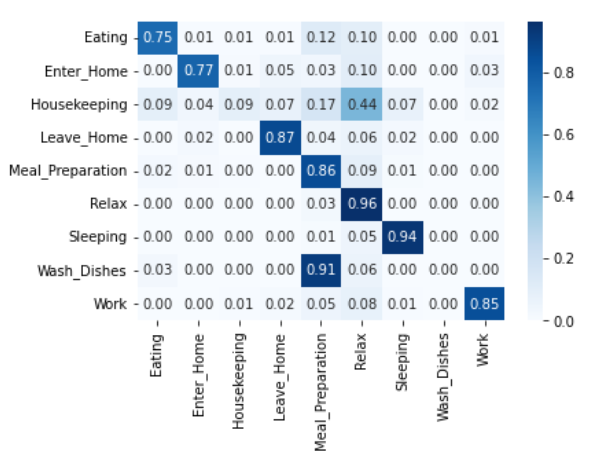} \\
    \small DeXAR & \small \acronym{}
  \end{tabular}
  \caption[Classification results on CASAS Aruba]{Confusion Matrices for CASAS Aruba.}
  \label{fig:cm_aruba}
\end{figure*}

\subsubsection{Explainability results}

As we previously mentioned, we leverage an LLM-based approach to compare \acronym{} and DeXAR. However, due to the costs of LLM-based APIs requests, we sampled $30$ random windows for each activity. Similarly to \cite{arrotta2022dexar}, we only evaluate the quality of explanations associated to correct predictions, since evaluating explanations of wrong predictions is still an open problem and we will consider it for future work.
For each window, we provide the LLM with the explanations generated by the two models and we obtain as output the explanation preferred by the LLM.
Table~\ref{tab:llm-example} shows an example of how the LLM compares the explanations.

\begin{table}[h]
    \caption{Example of LLM-based explaination evaluation.}
    \label{tab:llm-example}
    \centering
\begin{tabular}{|p{2.5cm}|p{9.5cm}|} 
  \hline 
  GNNExplainer & I predicted leaving home mainly due to the following observations:  Bob approached the area between the dining room and the hall multiple times, and then the front door. \\
  \hline
  DeXAR &  I predicted leaving home mainly because Bob has been near the front door and between the dining room and the hall.\\
  \hline
  LLM answer &  Let's break down each explanation: \\ & \\ & 1) The first explanation mentions specific observations of Bob approaching the area between the dining room and the hall multiple times before heading to the front door. This sequence of movements aligns with the common behavior of someone preparing to leave home, such as gathering belongings or checking for keys before exiting. Therefore, this explanation provides a clear and logical connection between Bob's movements and the prediction of leaving home. \\ & \\ & 2) The second explanation is similar to the first one but lacks the detail of multiple approaches to the specific areas. While it still mentions Bob being near the front door and between the dining room and the hall, the repetition of movements in the first explanation adds more weight to the prediction of leaving home.  \\ & \\ & Considering the clarity and specificity of the observations provided in the first explanation, it offers a more detailed and plausible rationale for predicting leaving home based on Bob's movements. \\ & \\ & CHOICE=1.\\
  \hline
\end{tabular}
\end{table}

Figure~\ref{fig:BP0} shows, for both datasets, the percentage of times where the LLM preferred explanations from \acronym{} compared to the ones generated by DeXAR. We observe that the explanations generated by \acronym{} were preferred by the LLM in $80\%$ of the times for the CASAS Milan dataset and $69\%$ of the times for the CASAS Aruba dataset.

\begin{figure}
\centering   \includegraphics[width=0.6\textwidth]{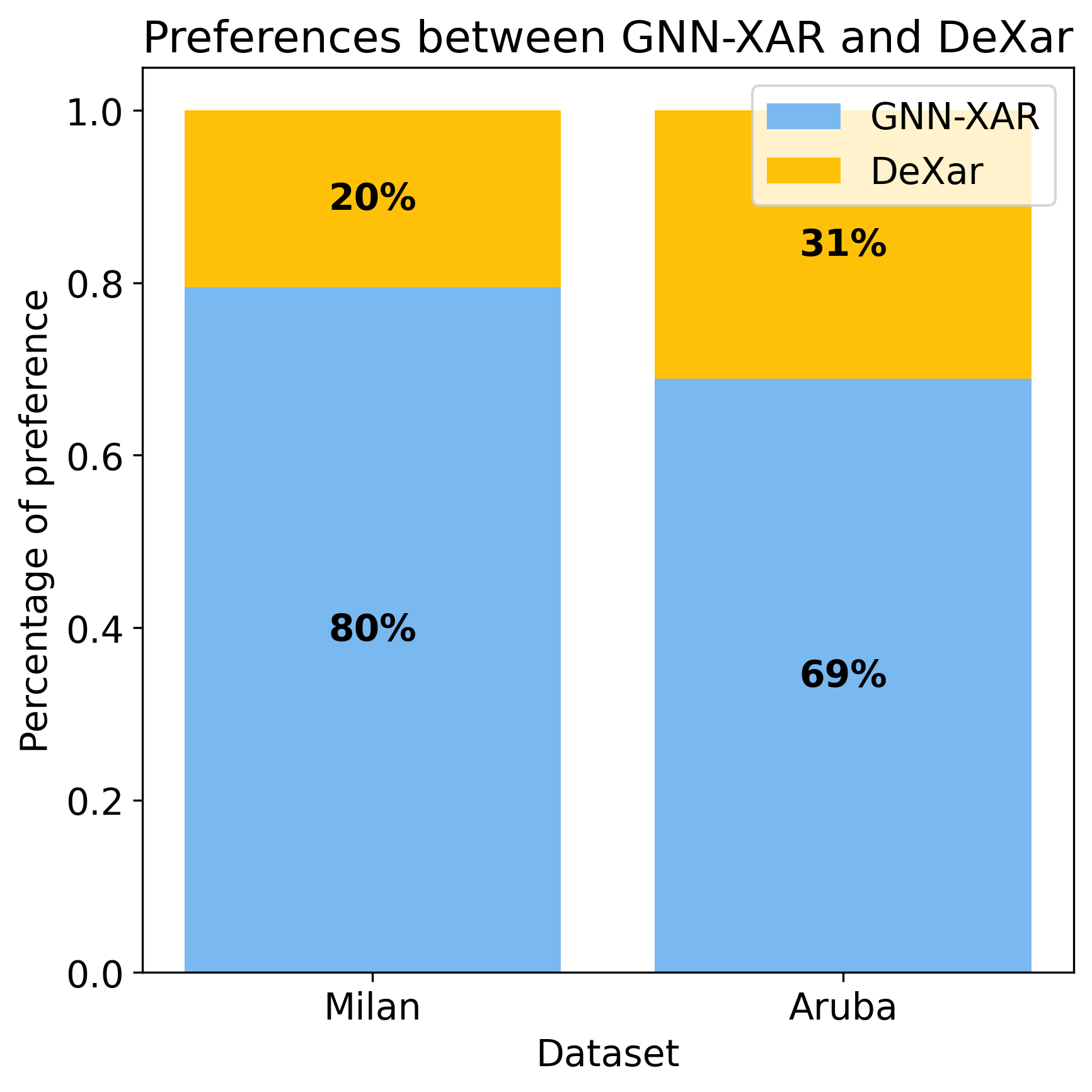}
    \caption[Explanation evaluation]{Overall percentage of preferences given by the LLM to explanations given by GNN-XAR and DeXar.}
    \label{fig:BP0}
\end{figure}

It is important to note that this preference is not uniform over all the classes classes. Indeed, Figures~\ref{fig:BP1} and~\ref{fig:BP2} shows that more dynamic activities, like entering and leaving home, eating, and preparing a meal achieve a higher score with respect to static activities like sleeping, reading, relaxing and watching TV.
This is reasonable since our graph encoding is better at capturing temporal relations in dynamic activities.
The only time where DeXAR explanations ``wins'' over the ones of \acronym{} is on the Sleep activity for the CASAS Aruba dataset. This is probably due to the fact that in this dataset there is a higher number of sensors in the bedroom, and a slight movement during sleep may trigger more sensors at once. Thus, the GNN may tend to explain the sequence of actions transmitting a false sense of movement from one sensor to the other. This probably can be fixed in future work by adding further heuristics to refine the explanations.

\begin{figure}
\centering   \includegraphics[width=0.8\textwidth]{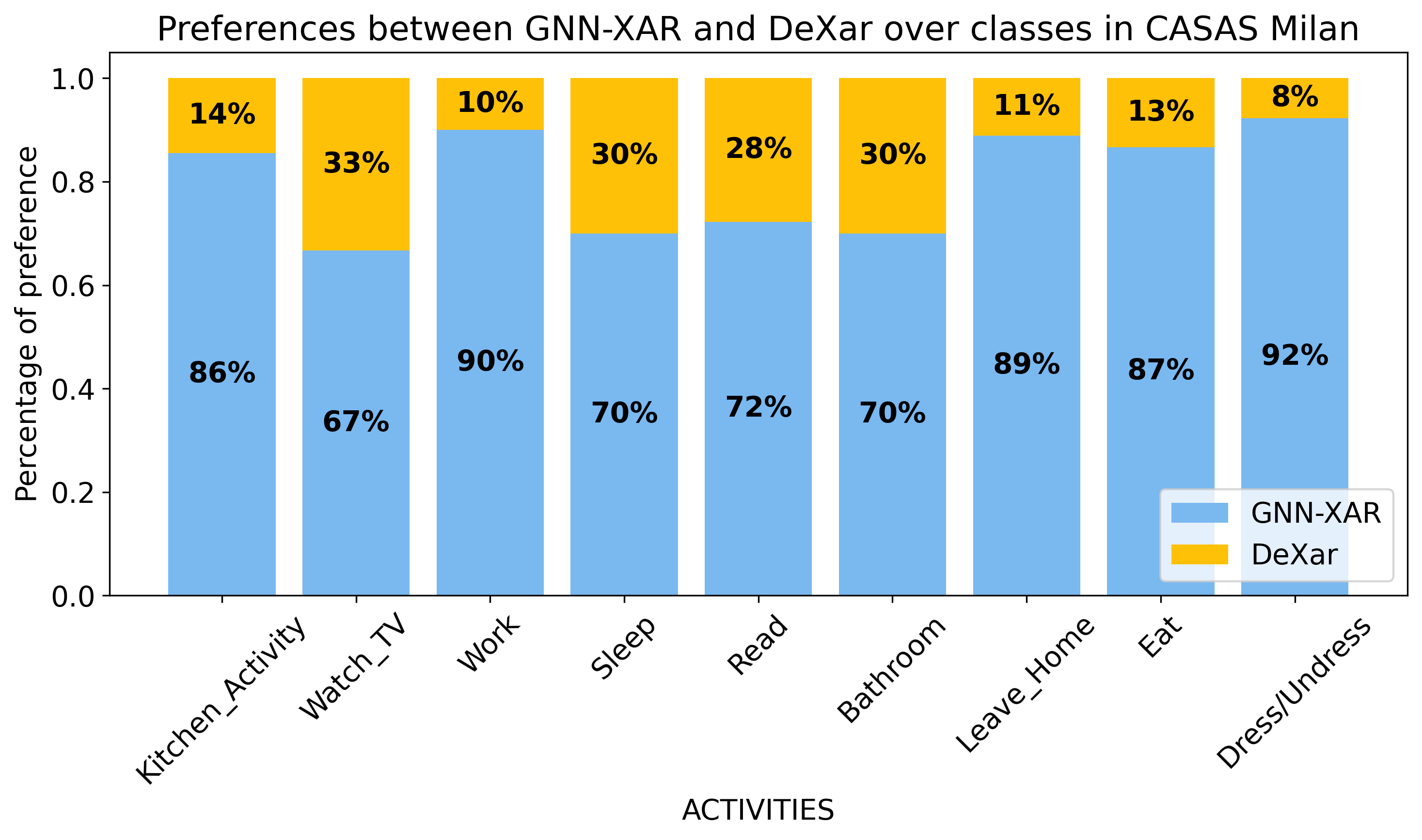}
    \caption[Explanation evaluation class by class on CASAS Milan]{Percentage of preferences (for each activity class) given by the LLM to explanations given by GNN-XAR and DeXar for CASAS Milan.}
    \label{fig:BP1}
\end{figure}

\begin{figure}
\centering   \includegraphics[width=0.8\textwidth]{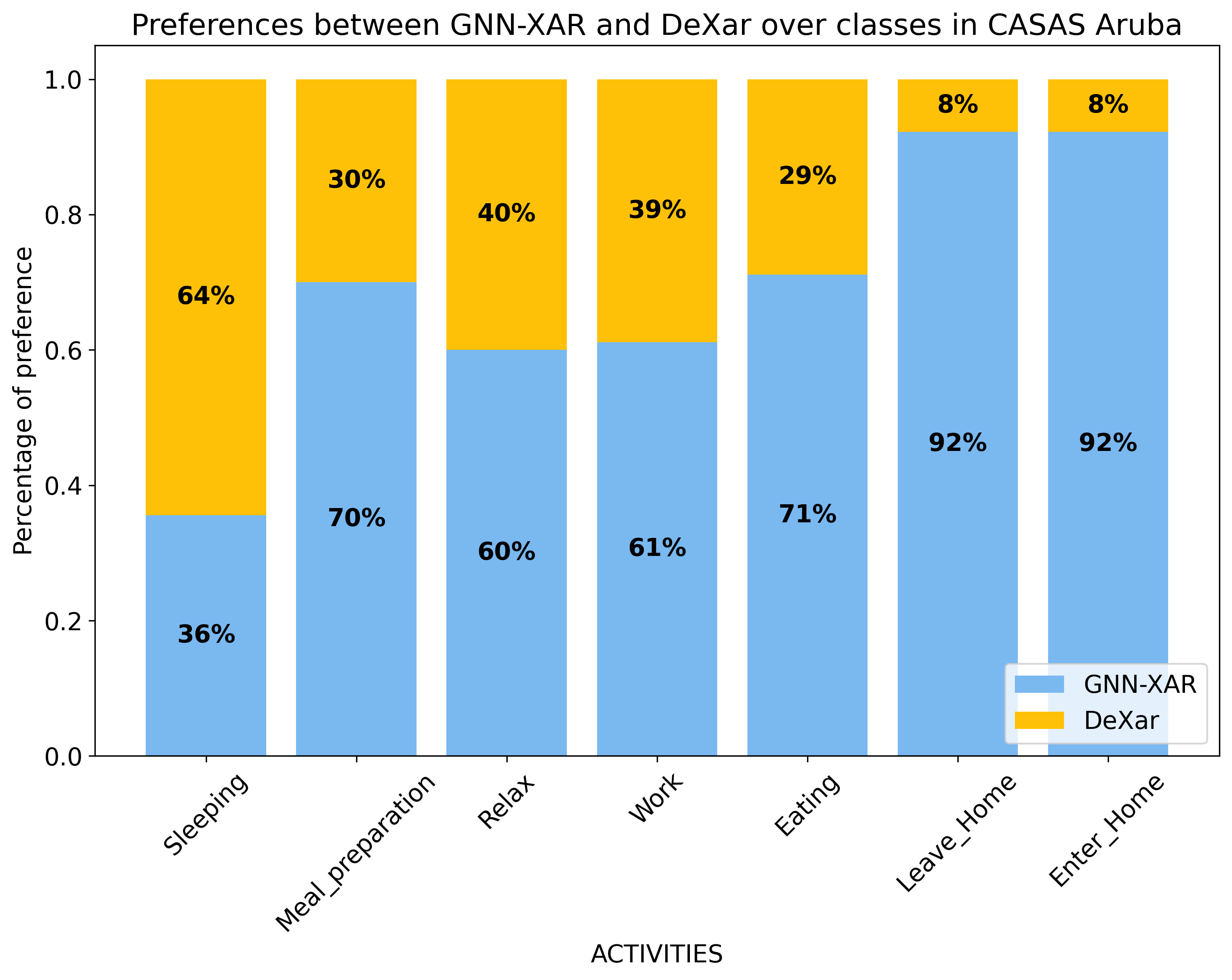}
    \caption[Explanation evaluation class by class on CASAS Aruba]{Percentage of preferences (for each activity class) given by the LLM to explanations given by GNN-XAR and DeXar for CASAS Aruba.}
    \label{fig:BP2}
\end{figure}

\section{Conclusion and Future Work}

In this paper we presented \acronym{}, an explainable Graph Neural Network framework for ADLs recognition in smart home environments.
Our results suggest that \acronym{} generates effective explanations by leveraging the structural properties of the graph representation.
While the results are promising, this work still has limitations, and we plan to extend it following several research directions.




Currently, \acronym{} only considers binary environmental sensors. We will investigate how to also integrate continuous sensor data from mobile/wearable devices, information about past activities, and other context information.

Another limitation of \acronym{} is that it assigns an importance score to a node or arc within a graph structure, but it does not consider nodes or arcs features in the explanations. Therefore, another possible development might consist in improving the explainer algorithm to also provide such details.




Regarding the segmentation, in this work we considered fixed time sliding windows, that is the standard approach. In future work, we will investigate the impact of dynamic segmentation~\cite{aminikhanghahi2017using} on \acronym{}.

Finally, while we used LLMs to evaluate the explanations, we will investigate where it is possible to leverage them also to automatically generate explanations starting from the most important nodes and arcs obtained by GNNexplainer.


\begin{credits}
\subsubsection{\ackname} 

This work was supported in part by MUSA, SERICS, and FAIR projects under the NRRP MUR program funded by the EU-NGEU. Views and opinions expressed are those of the authors only and do not necessarily reflect those of the European Union or the Italian MUR. Neither the European Union nor the Italian MUR can be held responsible for them.

\subsubsection{\discintname}
The authors have no competing interests to declare that are
relevant to the content of this article.
\end{credits}

\bibliographystyle{splncs04}
\bibliography{references}

\begin{thebibliography}{10}
\providecommand{\url}[1]{\texttt{#1}}
\providecommand{\urlprefix}{URL }
\providecommand{\doi}[1]{https://doi.org/#1}

\bibitem{agarwal2023evaluating}
Agarwal, C., Queen, O., Lakkaraju, H., Zitnik, M.: Evaluating explainability for graph neural networks. Scientific Data  \textbf{10}(1), ~144 (2023)

\bibitem{aminikhanghahi2017using}
Aminikhanghahi, S., Cook, D.J.: Using change point detection to automate daily activity segmentation. In: 2017 IEEE International Conference on Pervasive Computing and Communications Workshops (PerCom Workshops). pp. 262--267. IEEE (2017)

\bibitem{arrieta2020explainable}
Arrieta, A.B., D{\'\i}az-Rodr{\'\i}guez, N., Del~Ser, J., Bennetot, A., Tabik, S., Barbado, A., Garc{\'\i}a, S., Gil-L{\'o}pez, S., Molina, D., Benjamins, R., et~al.: Explainable artificial intelligence (xai): Concepts, taxonomies, opportunities and challenges toward responsible ai. Information Fusion  \textbf{58},  82--115 (2020)

\bibitem{arrotta2022dexar}
Arrotta, L., Civitarese, G., Bettini, C.: Dexar: Deep explainable sensor-based activity recognition in smart-home environments. Proceedings of the ACM on Interactive, Mobile, Wearable and Ubiquitous Technologies  \textbf{6}(1),  1--30 (2022)

\bibitem{atzmueller2018explicative}
Atzmueller, M., Hayat, N., Trojahn, M., Kroll, D.: Explicative human activity recognition using adaptive association rule-based classification. In: 2018 IEEE International Conference on Future IoT Technologies (Future IoT). pp.~1--6. IEEE (2018)

\bibitem{bettini2021explainable}
Bettini, C., Civitarese, G., Fiori, M.: Explainable activity recognition over interpretable models. In: 2021 {IEEE} International Conference on Pervasive Computing and Communications Workshops. pp. 32--37. IEEE (2021)

\bibitem{bouchabou2021survey}
Bouchabou, D., Nguyen, S.M., Lohr, C., LeDuc, B., Kanellos, I.: A survey of human activity recognition in smart homes based on iot sensors algorithms: Taxonomies, challenges, and opportunities with deep learning. Sensors  \textbf{21}(18), ~6037 (2021)

\bibitem{cook2010learning}
Cook, D.J.: Learning setting-generalized activity models for smart spaces. IEEE intelligent systems  \textbf{2010}(99), ~1 (2010)

\bibitem{cook2009assessing}
Cook, D.J., Schmitter-Edgecombe, M.: Assessing the quality of activities in a smart environment. Methods of information in medicine  \textbf{48}(05),  480--485 (2009)

\bibitem{das2023explainable}
Das, D., Nishimura, Y., Vivek, R.P., Takeda, N., Fish, S.T., Ploetz, T., Chernova, S.: Explainable activity recognition for smart home systems. ACM Transactions on Interactive Intelligent Systems  \textbf{13}(2),  1--39 (2023)

\bibitem{10.1145/3565973}
Dong, G., Tang, M., Wang, Z., Gao, J., Guo, S., Cai, L., Gutierrez, R., Campbel, B., Barnes, L.E., Boukhechba, M.: Graph neural networks in iot: A survey. ACM Trans. Sen. Netw.  \textbf{19}(2) (apr 2023)

\bibitem{duan2022multivariate}
Duan, Z., Xu, H., Wang, Y., Huang, Y., Ren, A., Xu, Z., Sun, Y., Wang, W.: Multivariate time-series classification with hierarchical variational graph pooling. Neural Networks  \textbf{154},  481--490 (2022)

\bibitem{fiori2024using}
Fiori, M., Civitarese, G., Bettini, C.: Using large language models to compare explainable models for smart home human activity recognition. In: Companion of the 2024 on ACM International Joint Conference on Pervasive and Ubiquitous Computing. pp. 881--884 (2024)

\bibitem{gu2021survey}
Gu, F., Chung, M.H., Chignell, M., Valaee, S., Zhou, B., Liu, X.: A survey on deep learning for human activity recognition. ACM Computing Surveys (CSUR)  \textbf{54}(8),  1--34 (2021)

\bibitem{guesgen2020using}
Guesgen, H.W.: Using rough sets to improve activity recognition based on sensor data. Sensors  \textbf{20}(6), ~1779 (2020)

\bibitem{jeyakumar2023x}
Jeyakumar, J.V., Sarker, A., Garcia, L.A., Srivastava, M.: X-char: A concept-based explainable complex human activity recognition model. Proceedings of the ACM on interactive, mobile, wearable and ubiquitous technologies  \textbf{7}(1),  1--28 (2023)

\bibitem{khodabandehloo2021healthxai}
Khodabandehloo, E., Riboni, D., Alimohammadi, A.: Healthxai: Collaborative and explainable ai for supporting early diagnosis of cognitive decline. Future Generation Computer Systems  \textbf{116},  168--189 (2021)

\bibitem{9995660}
Liao, T., Zhao, J., Liu, Y., Ivanov, K., Xiong, J., Yan, Y.: Deep transfer learning with graph neural network for sensor-based human activity recognition. In: 2022 IEEE International Conference on Bioinformatics and Biomedicine (BIBM). pp. 2445--2452 (2022)

\bibitem{liciotti2020sequential}
Liciotti, D., Bernardini, M., Romeo, L., Frontoni, E.: A sequential deep learning application for recognising human activities in smart homes. Neurocomputing  \textbf{396},  501--513 (2020)

\bibitem{meena2023explainable}
Meena, T., Sarawadekar, K.: An explainable self attention based spatial-temporal analysis for human activity recognition. IEEE Sensors Journal  (2023)

\bibitem{9767342}
Mohamed, A., Lejarza, F., Cahail, S., Claudel, C., Thomaz, E.: Har-gcnn: Deep graph cnns for human activity recognition from highly unlabeled mobile sensor data. In: 2022 IEEE International Conference on Pervasive Computing and Communications Workshops and other Affiliated Events (PerCom Workshops). pp. 335--340 (2022)

\bibitem{mondal2020new}
Mondal, R., Mukherjee, D., Singh, P.K., Bhateja, V., Sarkar, R.: A new framework for smartphone sensor-based human activity recognition using graph neural network. IEEE Sensors Journal  \textbf{21}(10),  11461--11468 (2020)

\bibitem{9874212}
Nian, A., Zhu, X., Xu, X., Huang, X., Wang, F., Zhao, Y.: Hgcnn: Deep graph convolutional network for sensor-based human activity recognition. In: 2022 8th International Conference on Big Data and Information Analytics (BigDIA). pp. 422--427 (2022)

\bibitem{ribeiro2016should}
Ribeiro, M.T., Singh, S., Guestrin, C.: " why should i trust you?" explaining the predictions of any classifier. In: Proceedings of the 22nd ACM SIGKDD international conference on knowledge discovery and data mining. pp. 1135--1144 (2016)

\bibitem{riboni2016smartfaber}
Riboni, D., Bettini, C., Civitarese, G., Janjua, Z.H., Helaoui, R.: Smartfaber: Recognizing fine-grained abnormal behaviors for early detection of mild cognitive impairment. Artificial intelligence in medicine  \textbf{67},  57--74 (2016)

\bibitem{9680185}
Sarkar, A., Sen, T., Roy, A.K.: Grafehty: Graph neural network using federated learning for human activity recognition. In: 2021 20th IEEE International Conference on Machine Learning and Applications (ICMLA). pp. 1124--1129 (2021)

\bibitem{s24123944}
Srivatsa, P., Plötz, T.: Using graphs to perform effective sensor-based human activity recognition in smart homes. Sensors  \textbf{24}(12) (2024)

\bibitem{10174597}
Wieland, C., Pankratius, V.: Tinygraphhar: Enhancing human activity recognition with graph neural networks. In: 2023 IEEE World AI IoT Congress (AIIoT). pp. 0047--0054 (2023)

\bibitem{wolf2019explainability}
Wolf, C.T.: Explainability scenarios: towards scenario-based xai design. In: Proceedings of the 24th International Conference on Intelligent User Interfaces. pp. 252--257 (2019)

\bibitem{ye2023graph}
Ye, J., Jiang, H., Zhong, J.: A graph-attention-based method for single-resident daily activity recognition in smart homes. Sensors  \textbf{23}(3), ~1626 (2023)

\bibitem{ying2019gnnexplainer}
Ying, Z., Bourgeois, D., You, J., Zitnik, M., Leskovec, J.: Gnnexplainer: Generating explanations for graph neural networks. Advances in neural information processing systems  \textbf{32} (2019)

\end{thebibliography}

\end{document}